\documentclass[letterpaper, 10 pt, conference]{ieeeconf}  

\IEEEoverridecommandlockouts                              

\usepackage{cite}
\usepackage{amsmath,amssymb,amsfonts}
\usepackage{algorithmic}
\usepackage{graphicx}
\usepackage{textcomp}
\usepackage{xcolor}

\makeatletter
\let\NAT@parse\undefined
\makeatother

\usepackage{hyperref}

\usepackage{multirow}
\usepackage{wrapfig}
\usepackage{makecell}

\usepackage{hyperref}
\hypersetup{
  colorlinks   = true,    
  urlcolor     = blue,    
  linkcolor    = blue,    
  citecolor    = blue      
}

\usepackage{dsfont}
\usepackage{pifont}

\newcommand{\methodname}{\texttt{ULT}}
\newcommand{\fullmethodname}{Unified Locomotion Transformer}

\begin{document}

\title{\LARGE \bf 
Unified Locomotion Transformer with Simultaneous Sim-to-Real Transfer for Quadrupeds
}

\author{Dikai Liu$^{1,2}$ \and Tianwei Zhang$^{2}$ \and Jianxiong Yin$^{1}$ \and Simon See$^{1,3}$
\thanks{$^{1}$ NVIDIA AI Technology Centre (NVAITC); e-mail: {\tt\small \{dikail,jianxiongy,ssee\}@nvidia.com}}
\thanks{$^{2}$ College of Computing and Data Science, Nanyang Technological University, Singapore; e-mail: {\tt\small dikai001@e.ntu.edu.sg, tianwei.zhang@ntu.edu.sg}}
\thanks{$^{3}$ also with Nanyang Technological University and Coventry University}
}

\maketitle

\begin{abstract}
Quadrupeds have gained rapid advancement in their capability of traversing across complex terrains. The adoption of deep Reinforcement Learning (RL), transformers and various knowledge transfer techniques can greatly reduce the sim-to-real gap. However, the classical teacher-student framework commonly used in existing locomotion policies requires a pre-trained teacher and leverages the privilege information to guide the student policy. With the implementation of large-scale models in robotics controllers, especially transformers-based ones, this knowledge distillation technique starts to show its weakness in efficiency, due to the requirement of multiple supervised stages. In this paper, we propose \fullmethodname{} (\methodname{}), a new transformer-based framework to unify the processes of knowledge transfer and policy optimization in a single network while still taking advantage of privilege information. The policies are optimized with reinforcement learning, next state-action prediction, and action imitation, all in just one training stage, to achieve zero-shot deployment. Evaluation results demonstrate that with \methodname{}, optimal teacher and student policies can be obtained at the same time, greatly easing the difficulty in knowledge transfer, even with complex transformer-based models. 
\end{abstract}


\section{Introduction}

Driven by the advancement of deep reinforcement learning (RL), quadruped robots have drawn great attention, due to their capability of traversing across complex terrains~\cite{tan2018sim,hwangbo2019learning,lee2020learning,kumar2021rma}. Traditional robotics controllers rely on dedicated models and heuristics, which require extensive prior knowledge and can struggle to adapt to dynamic environments and unpredictable situations.
Recently, a new paradigm is introduced to push quadrupeds' limit to handle complex tasks in challenging environments~\cite{lee2020learning,kumar2021rma,krotkov2018darpa,bellicoso2018advances,chen2021autonomous,hooks2020alphred}: a controller is learned from simulations that contain various environmental and physical factors, and then transferred to the physical robot through RL and various knowledge transfer techniques~\cite{yu2017preparing,lee2020learning,kumar2021rma,lai2023sim,radosavovic2023learning} to overcome the sim-to-real gap using privilege information.

One of the commonly used knowledge distillation methods is the teacher-student framework~\cite{lee2020learning}. A teacher policy is first trained through RL, and privilege information about the environment is provided for learning the optimal locomotion strategies efficiently. As additional information, including ground-truth dynamics and terrain profiles, is often inaccessible in the real world, a student policy is subsequently trained to make the model deployable. This is typically achieved through supervised knowledge distillation, either offline~\cite{lee2020learning} or online~\cite{kumar2021rma} fashion by creating a data set with new trajectories and associated teacher action labels through algorithms such as the Data Aggregator (DAgger)~\cite{ross2011reduction}. 

Sequential training with policy imitation for locomotion control is generally not data-efficient~\cite{ji2022concurrent}. The performance of the student is limited by the teacher policy and the robustness depends on the diversity of the data set. Generally, they will deteriorate if the situation deviates from the trajectories in real-world deployment. To address this, DreamWaQ~\cite{nahrendra2023dreamwaq} was introduced with the asymmetric actor-critic architecture~\cite{pinto2017asymmetric} and context estimation, including next observation, base velocity and latent space. By concurrently training with proprioceptive observation for exploration and privileged information as the critic, it allows the agent to explore with indirect guidance. 

Recently, self-attention-based transformers~\cite{vaswani2017attention} have been widely introduced into robotics in both lower control (e.g., legged locomotion directly~\cite{lai2023sim,radosavovic2023learning,radosavovic2024humanoid,liu2024masked} or with a command interface~\cite{tang2023saytap}) and high-level decision making with multimodal processes~\cite{reed2022generalist, brohan2022rt, brohan2023rt,padalkar2023open} with their capability to handle variable context lengths, sensor combinations~\cite{liu2024masked} and even robot embodiments~\cite{sferrazza2024body}. In direct locomotion controllers, transformers have demonstrated superior capability in temporal information capture compared to recurring neural networks (RNN) and temporal convolutional networks (TCN). However, training transformers is data-hungry in general. For example, when combing the vanilla transformer with legged locomotion, Lai et al.~\cite{lai2023sim} generated an additional 40M timesteps for 400K updates using a two-stage supervised training approach. Similarly, Radosavovic et al.~\cite{radosavovic2023learning} doubled the number of timesteps for joint supervised transfer compared to the teacher policy training stage. These approaches are not only time-consuming, but also necessitating additional setup to handle multiple models and the large volumes of generated data.

Motivated by the above limitation, the objective of this paper is to simplify the knowledge distillation process with a unified architecture to keep teacher and student policies in a single network for simultaneous optimization. Our inspiration comes from the exceptional capabilities of transformers in multimodal modeling of temporal and sensory information and context understanding~\cite{radosavovic2024humanoid}, which can well support policy optimization~\cite{nahrendra2023dreamwaq}. Thus, we can introduce the privilege information into the observation as another modality to form a unified framework for single-phase optimization of teacher and student policies, achieving zero-shot sim-real transfer. 

To this end, we propose \fullmethodname{} (\methodname{}), a new unified framework for end-to-end quadruped locomotion. It is based on the standard transformer architecture with casual masking to pack teacher and student policies in a single network. These policies are optimized jointly with reinforcement learning, next state-action prediction, and action imitation, all in just one training phase, to overcome the sim-to-real gap and achieve zero-shot deployment. In this way, we eliminate the need for dedicated design and training of a teacher network, while the privilege information can still efficiently guide the student policy during the exploration to generate more diverse trajectories to improve the overall generalization and robustness. 

We extensively evaluate our framework in simulation and compare it with state-of-the-art knowledge transfer baselines. We also deploy it directly in the real world for practicality validation. Evaluation results demonstrate that \methodname{} exhibits better performance with less trajectory information needed, indicating its higher efficiency with the help from next state-action predication, action imitation and mixed exploration. With unified training in one single phase for simultaneous teacher and student policy optimization, we ease the pipeline of knowledge distillation for zero-shot sim-to-real transfer.

\section{Related Work}

\subsection{Knowledge Transfer in RL-based Legged Locomotion}
With the increased attention in Reinforcement Learning (RL) and the advance of robotics simulation, the RL-based simulation-first controller has been dominating legged locomotion~\cite{tan2018sim,hwangbo2019learning,lee2020learning,kumar2021rma,lai2023sim}, eliminating the need for extensive prior knowledge and reducing the cost of training by massive parallelism~\cite{rudin2022learning}. To transfer the policy from simulation to the physical world, Yu et al.~\cite{yu2017preparing} used online system identification to infer physics parameters in the real world. To further advance the locomotion policy for optimized control and close the sim-to-real gap, Lee et al.~\cite{lee2020learning} introduced a teacher-student framework for action cloning, which uses historical proprioceptive data to infer teacher behaviors leveraging the privilege information. This framework has been widely adopted in subsequent works. To improve the trajectory generation during transfer, Kumar et al.~\cite{kumar2021rma} used a randomly initialized student policy for online adaptation with better exploration to improve the robustness. The introduction of Transformer-based controllers in legged locomotion makes the optimization more challenging. Lai et al.~\cite{lai2023sim} introduced a two-stage transfer to ensure that the student policy can gather useful trajectories during online correlation for fast convergence. Radosavovic et al.~\cite{radosavovic2023learning} combined RL exploration and teacher supervision to jointly optimize the student. 

\subsection{Transformer in Legged Locomotion}

Transformer-based models have recently made significant inroads into robotics with their exceptional capabilities for in-context understanding and human-robot interaction. They are equipped with Large Vision Language Models (VLM)~\cite{brohan2022rt} and Vision Language Action Models (VLA)~\cite{brohan2023rt,padalkar2023open,kim2024openvla}, to handle the multimodal input from humans and physical world. 

Constrained by the computational capability and power consumption of real-time inference on the edge devices, the transformer used for legged locomotion is often much smaller and sometimes a high-level command is adopted
as the control interface. Yang et al.~\cite{yang2021learning} developed a transformer model for vision-based locomotion, which outputs desired velocity commands for a dedicated low-level controller to conduct motor output. Similarly, Tang et al.~\cite{tang2023saytap} used the gait pattern as the command interface for quadruped locomotion. With knowledge transfer techniques, Lai et al.~\cite{lai2023sim} and Radosavovic et al.~\cite{radosavovic2023learning} utilized historical observation-action information for direct motor command in quadruped and bipedal locomotion, respectively. Radosavovic et al.~\cite{radosavovic2024humanoid} further expanded the framework with next token prediction to utilize different data source including captured real-world locomotion trajectories. 

To improve the generalization of transformer-based locomotion controllers, different masking strategies are used in recent works to partially remove information during training and testing so the controller can learn to focus on the most important information. Sferrazza et al.~\cite{sferrazza2024body} used body-induced bias based on the embodiment graph for an embodiment-aware transformer. Liu et al.~\cite{liu2024masked} introduced masking directly at the sensor level for generalization with different sensor combinations as input. Transformer-based controllers can also achieve cross-embodiment policy to conduct different types of tasks on various robots with one single network~\cite{bohlinger2024one,doshi2024scaling}

\section{Simulation Environment}

In this paper, we implement the simulation environment in Isaac Gym~\cite{makoviychuk2021isaac} and IsaacGymEnvs~\cite{makoviychuk2021isaac} to train locomotion agents in large scale parallelism. All baselines and variants of our method are trained with the exact same simulation setups to ensure fair comparison. 

\subsection{Terrain and Curriculum}
To ensure that the policy is robust against different indoor and outdoor environments, we adopt the terrain curriculum from~\cite{rudin2022learning} with smooth slope, rough slope, stairs up, stairs down and discrete obstacle terrains. Each type of terrain has 10 levels with incremental difficulty and an overall proportion of $[0.1, 0.1, 0.35, 0.25, 0.2]$, respectively. The linear velocity return is tracked across each trajectory's life cycle. The level is considered solved when an agent reaches 80\% of the maximum achievable tracking reward and progresses to the next level. If any agent fails to reach 25\% of the maximum reward, it will regress to a lower level. 

\subsection{Domain Randomization}
\label{sec:dr}
Following ~\cite{rudin2022learning,nahrendra2023dreamwaq}, we apply Domain Randomization (DR) on key dynamics parameters to enhance the robustness of the policy. To simulate the actual user commands, we sample the linear commands in longitudinal and lateral direction separately with a uniform distribution in $[-0.5, 0.5]$ m/s. For the angular command, we first sample the desired heading of the robot and cap the resulted angular velocity command at 0.5 rad/s. The commands are resampled every 10 seconds. Table~\ref{tab:dr} lists the key parameters of DR used in the simulation environment. 

\begin{table}[t]
    \centering
    \caption{Simulation parameters of domain randomization.}
    \label{tab:dr}
    \begin{tabular}{ccc}
        \Xhline{1.5pt}
        \textbf{Parameters}  & \textbf{Range} & \textbf{Unit} \\ \Xhline{1.5pt}
        Linear Command       & [-0.5, 0.5]        & $m/s$           \\
        Angular Heading      & [-3.14, 3.14]        & $rad$         \\ \hline
        $K_p$ Scale          & [0.9, 1.1]     & -             \\
        $K_d$ Scale          & [0.9, 1.1]     & -             \\
        Friction Scale       & [0.7, 1.3]     & -             \\
        Motor Strength Scale & [0.9. 1.1]     & -             \\
        Payload              & [0, 5]         & $kg$            \\
        Payload CoM Offset   & [-0.1, 0.1]    & $m$             \\
        External Push        & [-1, 1]        & $m/s$           \\
        Gravity              & [9,41, 10.21]  & $m/s^2$       \\
        System Delay         & [0, 0.015]     & $s$             \\ \Xhline{1.5pt}
    \end{tabular}
\end{table}

\subsection{Observations and Actions}
\noindent\textbf{Privilege Information.} To achieve an optimal locomotion policy with the hidden information about the environment, related privilege data is extracted from simulation to form the privilege observation $e_t$ for the teacher policy to utilize. The privilege information contains randomized dynamics parameters $d_t$ sampled from Sec.~\ref{sec:dr}, ground truth robot states $s_t$ including base velocity, orientation, and precise surrounding height map $m_t$. Although such information is often inaccessible in the real world, it helps the teacher policy reconstruct states and improve learning efficiency~\cite{lee2020learning,kumar2021rma}.

\noindent\textbf{Proprioceptive Observation.} In order to conduct knowledge transfer for a deployable agent, the student policy only relies on onboard sensors to provide observations. Typically, quadrupedal robots are equipped with multiple sensors, including joint encoders, IMU, and foot contact sensors, which can provide information on joint position $q \in \mathbb{R}^{12}$, joint velocity $\dot{q} \in \mathbb{R}^{12}$, angular velocity $\omega \in \mathbb{R}^{3}$, gravity vector $g \in \mathbb{R}^{3}$ and binary foot contact $c \in \mathbb{R}^{4}$. In order to follow the user command, the agent also needs access to the randomly sampled $\mathrm{cmd} = [v_x, v_y, \omega_z] \in \mathbb{R}^{3}$ to form the observation of each step $o_t = [q, \dot{q}, \omega, g, c, \mathrm{cmd}] \in \mathbb{R}^{37}$. To provide the state transition and temporal information, the actions of the previous step $a_{t-1} \in \mathbb{R}^{12}$ are added with a list of historical information $\mathcal{T} = [a_0, o_1, a_1, o_2, \cdots, a_{t-1}, o_{t}]$. We use a rolling window of $t = 15$, resulting in a full observation in the space of $\mathbb{R}^{49 \times 15}$.

\noindent\textbf{Actions.} As typical RL locomotion policies perform inference at the frequency of 50-100 Hz, both the teacher and student policies output the desired joint position $q=a_t$, which is passed to a PD controller running at a much higher frequency for a smooth torque output:
\begin{equation}
    \tau = K_p (\hat{q}-q) + K_d (\hat{\dot q}- \dot q)
\end{equation}
with base stiffness $K_p$ and damping $K_d$ set to 30 and 0.7, respectively, and additional DR is added on. The target joint velocity $\hat{\dot q}$ is set to 0.

\subsection{Reward Function}
We follow the classic reward function design for omni-direction locomotion~\cite{rudin2022learning,kumar2021rma,nahrendra2023dreamwaq} to encourage the agent to follow the commanded velocity and primarily penalize the linear and angular movement along other axes, large joint acceleration and excessive power consumption.
The complete reward structure is detailed in Table~\ref{tab:reward}.
\begin{table}[t]
    \centering
    \caption{Reward terms for reinforcement learning}
    \label{tab:reward}
    \begin{tabular}{ccc}
        \Xhline{1.5pt}
        \textbf{Reward}           & \textbf{Definition}                         & \textbf{Scale} \\ \Xhline{1.5pt}
        Linear Velocity Tracking  & $\exp{(-5\|v_{xy}^{\mathrm{cmd}} - v_{xy}\|^2)}$         & 1.0            \\
        Angular Velocity Tracking & $\exp{(-5(\omega_z^{\mathrm{cmd}} - \omega_z)^2)}$     & 0.5            \\ \hline
        Body Z Velocity           & $\| v_z\|^2$                                & -2.0           \\
        Body Rotation             & $\| \omega_z\|^2$                           & -0.05          \\
        Joint Acceleration        & $\|\ddot{\theta}\|^2$                       & -2.5e-7        \\
        Output Work               & $\int \| \tau \cdot \dot{\theta} \|$        & -2.e-5         \\
        Action Rate               & $(a_t - a_{t-1})^2$                         & -0.05          \\
        Feet Slip                 & $\|g_t \cdot v_{xy}^{feet}\|$               & -0.1           \\
        Collision                 & $\mathds{1}_{collision}$                    & -1             \\ \Xhline{1.5pt}
    \end{tabular}
\end{table}

\section{Methodology}

\begin{figure*}[ht]
    \centering
    \includegraphics[width=.85\textwidth]{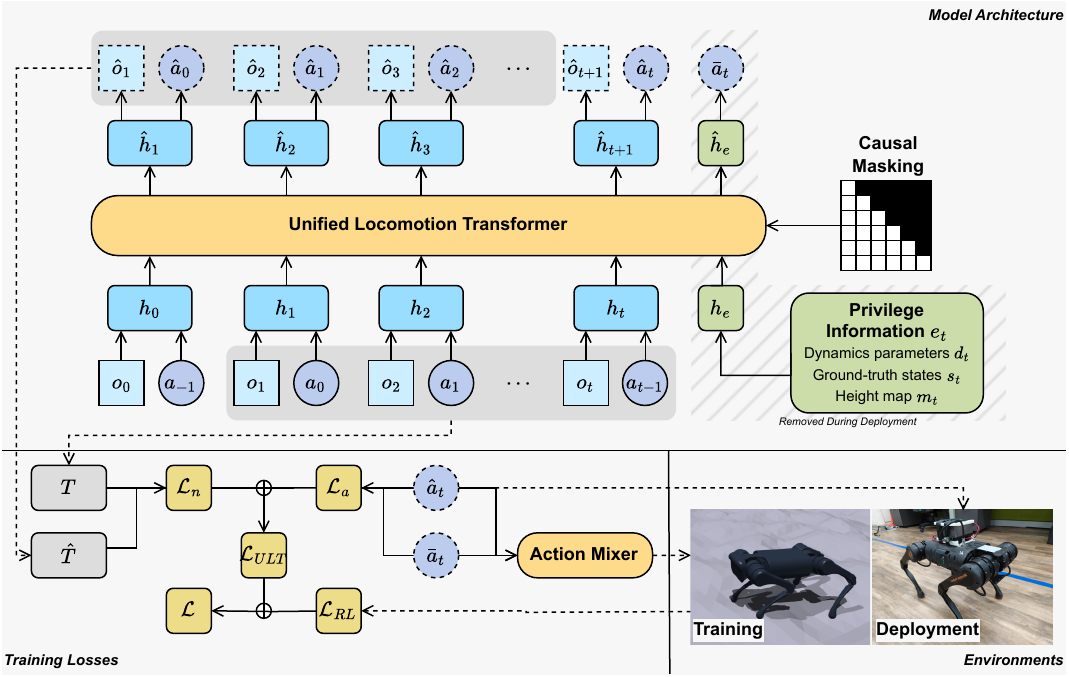}
    \caption{Illustration of the \fullmethodname{} (\methodname{}) framework. \methodname{} is a vanilla transformer-based architecture to unify the optimization of locomotion policy and knowledge transfer. With state-action trajectories and privilege information in a single framework, both teacher and student actions can be generated simultaneously. The optimization is conducted jointly through PPO by combining the RL loss and transformer loss, which contains the next state-action prediction for future trajectories, and action imitation between student and teacher policies. During training in simulation, an action mixer is used to ensure both policies are played to enhance exploration.  During the physical deployment, only proprioceptive observation is used for student actions to achieve zero-shot sim-to-real transfer.}
    \label{fig:overview}
    \vspace{-10pt}
\end{figure*}

We present \methodname{}, a transformer-based framework to unify the process of knowledge transfer and policy optimization in a single network leveraging the privilege information. Compared to the classic teacher-student transfer solution that requires a pre-trained teacher policy~\cite{lee2020learning,kumar2021rma,lai2023sim}, \methodname{} optimizes both policies jointly in a single phase to simplify the sim-to-real pipeline and reduce the total number of generated trajectories. Figure \ref{fig:overview} shows an overview of \methodname{}.

RL-based quadrupedal locomotion is often formulated as a Partially Observable Markov Decision Process (POMDP), defined by a tuple $\mathcal{M} = (\mathcal{S}, \mathcal{A}, T, \mathcal{R}, \Omega, \mathcal{O}, \gamma)$, where $\mathcal{S}, \mathcal{A}, \mathcal{R}, \Omega$ are the state, action, reward and observation spaces, respectively. The ground truth states $s_t \in \mathcal{S}$ give the most important information about the environment and can be accessed by the teacher policy in the simulation to search for an optimal control policy $\pi^*(a_{t+1} | s_t)$ by maximizing the sum of discounted future rewards:
\begin{equation}
    \pi^*(s, a) := \arg\max_{\pi} \mathbb{E}_{s_{t+1} \sim T(\cdot | s_t, a_t)} \left[ \sum_{t=0}^{\infty} \gamma^t r(s_t, a_t) \right]
\end{equation}
However, such information is often inaccessible in the real world, and the student policy can only rely on noisy proprioceptive observation $o_t \in  \Omega$ during deployment. Essentially, knowledge transfer aims to find the observation probabilities $\mathcal{O}: \mathcal{S}  \rightarrow \Omega$ such that the student policy can estimate its current status for decision making, either with latent space estimation~\cite{kumar2021rma} or direct action imitation~\cite{lee2020learning,lai2023sim,radosavovic2023learning}.

Although many existing quadruped locomotion solutions have used the historical information $\mathcal{T}$ to capture the temporal information~\cite{lee2020learning,kumar2021rma,lai2023sim}, they only treat it as a whole when predicting the next action $a_{t+1} = \pi(\mathcal{T})$ while missing the transition information $ T(s_{t+1} | s_t, a_t)$ hidden in the sequence itself. With the transformer architecture and its next token prediction capability~\cite{radosavovic2024humanoid}, we can extract and utilize information about the state-action transition from the history $\mathcal{T}$. By pairing it with policy imitation from the privilege information and mixed exploration, we can greatly improve the sample efficiency to discovery the state transition and observation probabilities, thus achieving the optimal teacher and student policies simultaneously in just one training phase.

\subsection{Model Architecture}

The foundational part of \methodname{} is a vanilla transformer~\cite{vaswani2017attention}. It contains multiple stacked multihead attention blocks with causal masking, so the tokens can only attend to themselves and the past tokens while the proprioceptive tokens will not access the information from the privilege tokens.

Similarly to~\cite{radosavovic2024humanoid}, we first tokenize the input trajectory $\mathcal{T}$ with a concatenated states action pair by a shared linear projection layer $W \in \mathbb{R}^{d \times (m+n)}$, where $m=37$ and $n=12$ are the dimensions of observation $o$ and $a$ at each step $t$. We choose $d=128$ as the size of the token embedding:
\begin{align} 
    z_t &= \mathrm{concat}(o_t, a_{t-1}) \\
    h_t &= W z_t 
\end{align} 

For privilege information $e_t = [d_t, s_t, m_t]$, we use an environmental factor encoder $\mu$ with a three-layer MLP similar to~\cite{kumar2021rma} to project it into the same embedding space:
\begin{equation}
    h_e = \mu(e)
\end{equation}

The transformer module takes the entire sequence of $H=[h_0, h_1, \cdots, h_t, h_e]$ with privilege information at the end to ensure that the information will not be leaked to the proprioceptive observation. The sequence is then processed through all the attention layers:
\begin{align}
    \begin{split}
        \hat{H} &= \mathrm{\methodname{}}(H) \\
                &= [\hat{h}_0, \hat{h}_1, \cdots, \hat{h}_t, \hat{h}_e] \label{eq:ult}
    \end{split}
\end{align}

\noindent\textbf{Next State-Action Prediction.} To perform an aligned prediction, we extract the first $(t-1)$ tokens from the output $\hat{H}_{0:(t-1)} = [\hat{h}_0, \hat{h}_1, \cdots, \hat{h}_{t-1}]$, and decode them through another shared linear project $\hat{W} \in \mathbb{R}^{(m+n) \times d}$ to predict the future state action trajectory for each step:
\begin{align} 
    \hat{z}_{t+1} &= \hat{W} \hat{h}_{t} \label{eq:last_decode} \\ 
    \hat{T} &= [\hat{z}_1, \hat{z}_2, \cdots, \hat{z}_t] 
\end{align}
We can compare the next state-action pairs between the predicted trajectory $\hat{T}$ and actual trajectory $T = [z_1, z_2, \cdots, z_t]$:
\begin{align} 
    \mathcal{L}_n &= \frac{1}{t} \sum_1^t \| z_t - \hat{z}_t \| ^2 \label{eq:L_n}
\end{align}

By optimizing $\mathcal{L}_n$, the transformer can always learn the transition relation of the robot state and action, regardless of the actions taken or the quality of the trajectories. 

\noindent\textbf{Action Output.} \methodname{} can simultaneously output actions from the teacher and student. For the student, with the next state-action prediction, we already implement a decoder layer to extract the predicted next state-action trajectory and it can already output the next action at each time step. Thus, we directly reuse the information in Eq.~\ref{eq:last_decode} to form the last concatenated state-action input token:
\begin{align}
    \hat{a}_t = (\hat{z}_{t+1})_{m:(m+n)}
\end{align}

For the teacher, with the process through all the attention layers, the resulted $\hat{h}_e$ (Eq.~\ref{eq:ult}) have already gathered all the information from the state-action trajectory due to casual masking. In order to generate actions from $\hat{h}_e$, we implement a policy $\pi$ with an MLP network similar to~\cite{kumar2021rma}:
\begin{equation}
    \bar{a}_t = \pi(\hat{h}_e)
\end{equation}

Thus, we can combine the imitation loss of the action and the next state-action prediction to get the overall performance with a weighting factor $\beta$:
\begin{align}
    \mathcal{L}_a &= \| \bar{a}_t - \hat{a}_t \| ^2 \label{eq.L_a} \\
    \mathcal{L}_{\mathrm{\methodname{}}} &= \mathcal{L}_n + \beta \mathcal{L}_a
\end{align}

\begin{table}[t]
    \centering
    \caption{Hyperparameters for PPO}
    \label{tab:hyper}
    \begin{tabular}{lr}
        \Xhline{1.5pt}
        \textbf{Parameters}   & \textbf{Value} \\ \Xhline{1.5pt}
        Number of GPUs        & 2              \\
        Actors per GPU        & 4096           \\
        Episode Length        & 20s            \\
        Horizon Length        & 24             \\
        Mini Epochs           & 5              \\
        Minibatch Size        & 16384          \\
        Learning Rate         & 3e-3           \\
        Scheduler             & cosine         \\
        Optimizer             & AdamW          \\
        Clip range            & 0.2            \\
        Entropy coefficient   & 0.005          \\
        Reward Discount       & 0.99           \\
        GAE Discount          & 0.95           \\
        Desired KL-divergence & 0.008          \\
        Weight Decay          & 0.01           \\
        \Xhline{1.5pt}
    \end{tabular}
\end{table}

\subsection{Action Mixer and Unified Training}
\label{sec:action_mixer}

In order to achieve optimized performance of $\bar{a}_t$ and $\hat{a}_t$ simultaneously without a pre-trained policy, online trajectories generated by both actions are needed. In a massive parallelism training environment with $X$ agents, an agent mask $M$ is created with a threshold $\alpha$ as the mix ratio such that:
\begin{align}
    \begin{split}
        a_i &= \begin{cases}
        \bar{a}_i, \text{if } M_i < \alpha \\
        \hat{a}_i, \text{otherwise}
        \end{cases} 
        \text{where } M_i \sim \mathcal{U}(0,1), i = 1,\cdots,X 
    \end{split}
\end{align}

Thus, a higher mix ratio $\alpha$ means more involvement of the teacher. The agent mask $M$ is frequently resampled to ensure exploration for both policies. With the trajectories generated by the resulted $a$, we use PPO~\cite{schulman2017proximal} to jointly optimize the policy and action imitation by appending the PPO RL loss and transformer loss with the hyperparameters in Table~\ref{tab:hyper}:
\begin{equation}
    \mathcal{L} = \mathcal{L}_{\mathrm{RL}} + \lambda \mathcal{L}_{\mathrm{\methodname{}}} 
\end{equation}

Optimization of $\mathcal{L}$ can simultaneously leverage the state transition, action imitation, and guidance from the teacher policy based on the privilege information to achieve a unified training in a single phase.

\subsection{Direct Sim-to-Real Deployment}
\label{sec:deploy}
The transformer architecture gives flexibility in variable input length. During training, causal masking ensures that future information will not be seen by previous tokens and the privilege information will never be leaked to proprioceptive observation. Thus, when deployed in the real world, we can directly remove the last privileged environmental token $h_e$ safely from the input sequence to generate the student action $\hat{a}_i$ directly with only the historical information $\mathcal{T}$.

\begin{figure*}[ht]
    \centering
    \includegraphics[width=\textwidth]{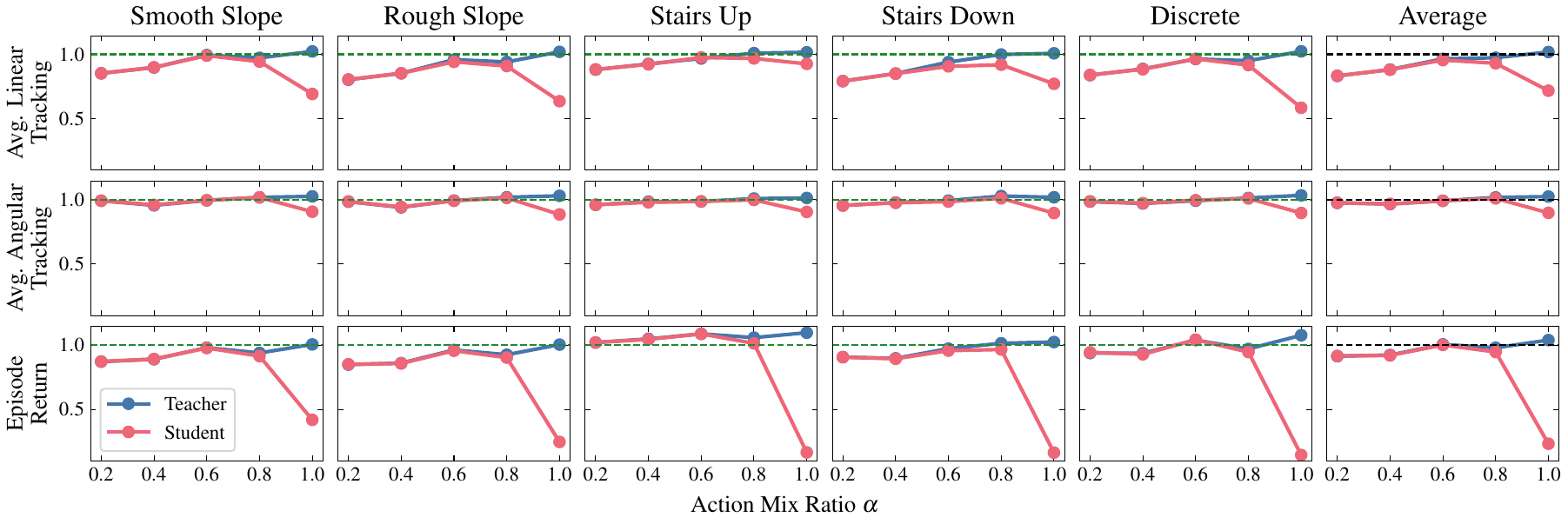}
    \caption{Performance of \methodname{} with different values of Action Mixer ratio $\alpha$ on five terrains and the overall performance across all trails.}
    \label{fig:mix_ratio}
    \vspace{-10pt}
\end{figure*}

\begin{table*}[ht]
\centering
\caption{Normalized Performance in key metrics with different baselines on five terrain types and average across all trails.}
\label{tab:comparison}
\resizebox{\textwidth}{!}{
\begin{tabular}{cccccccccccc}
\Xhline{1.5pt}
\multirow{2}{*}{\textbf{Terrain}} & \multirow{2}{*}{\textbf{Metric}} & \multicolumn{2}{c}{\textbf{ULT (Ours)}} & \textbf{} & \multicolumn{3}{c}{\textbf{Supervised Transfer}}                   & \textbf{} & \multirow{2}{*}{\textbf{\begin{tabular}[c]{@{}c@{}}Joint\\      Transfer\end{tabular}}} & \multirow{2}{*}{\textbf{CENet}} & \multirow{2}{*}{\textbf{PPO}} \\ \cline{3-4} \cline{6-8}
                                  &                                  & \textbf{Teacher}   & \textbf{Student}   & \textbf{} & \textbf{Offline-Only} & \textbf{Online-Only} & \textbf{Two-Stages} &           &                                                                                         &                                 &                               \\ \Xhline{1.5pt} 
\multirow{3}{*}{Smooth Slope} & Avg. Linear Tracking    & 0.994          & 0.990         &  & 0.691        & 0.994       & 1.003      &  & 0.907                                                                          & 1.005                  & 0.956                \\
                              & Avg. Angular Tracking   & 0.995          & 0.995         &  & 0.851        & 1.006       & 1.002      &  & 0.958                                                                          & 1.020                  & 0.977                \\
                              & Total Episode Return    & 0.980          & 0.978         &  & 0.412        & 0.999       & 0.993      &  & 0.880                                                                          & 1.001                  & 0.916                \\ \hline
\multirow{3}{*}{Rough Slope}  & Avg. Linear Tracking    & 0.959          & 0.941         &  & 0.660        & 0.985       & 0.990      &  & 0.882                                                                          & 0.950                  & 0.926                \\
                              & Avg. Angular Tracking   & 0.993          & 0.992         &  & 0.844        & 1.007       & 0.997      &  & 0.957                                                                          & 1.015                  & 0.987                \\
                              & Total Episode Return    & 0.962          & 0.957         &  & 0.432        & 0.977       & 0.967      &  & 0.876                                                                          & 0.965                  & 0.911                \\ \hline
\multirow{3}{*}{Stairs Up}    & Avg. Linear Tracking    & 0.967          & 0.965         &  & 0.837        & 0.962       & 0.995      &  & 0.943                                                                          & 0.966                  & 0.933                \\
                              & Avg. Angular Tracking   & 0.987          & 0.985         &  & 0.844        & 1.000       & 0.973      &  & 0.965                                                                          & 0.982                  & 0.982                \\
                              & Total Episode Return    & 1.088          & 1.086         &  & 0.138        & 0.829       & 0.824      &  & 0.890                                                                          & 1.074                  & 0.933                \\ \hline
\multirow{3}{*}{Stairs Down}  & Avg. Linear Tracking    & 0.939          & 0.906         &  & 0.705        & 0.948       & 0.953      &  & 0.884                                                                          & 0.872                  & 0.890                \\
                              & Avg. Angular Tracking   & 0.993          & 0.986         &  & 0.837        & 0.993       & 0.979      &  & 0.956                                                                          & 0.961                  & 0.967                \\
                              & Total Episode Return    & 0.973          & 0.957         &  & 0.125        & 0.836       & 0.879      &  & 0.923                                                                          & 0.929                  & 0.877                \\ \hline
\multirow{3}{*}{Discrete}     & Avg. Linear Tracking    & 0.964          & 0.964         &  & 0.613        & 1.000       & 0.996      &  & 0.848                                                                          & 0.915                  & 0.857                \\
                              & Avg. Angular Tracking   & 0.992          & 0.997         &  & 0.805        & 1.012       & 1.003      &  & 0.944                                                                          & 0.983                  & 0.957                \\
                              & Total Episode Return    & 1.039          & 1.043         &  & 0.107        & 1.009       & 0.990      &  & 0.889                                                                          & 0.926                  & 0.905                \\ \Xhline{1.5pt} 
\multirow{3}{*}{Average}      & Avg. Linear Tracking    & 0.965          & 0.953         &  & 0.698        & 0.978       & 0.987      &  & 0.892                                                                          & 0.942                  & 0.912                \\
                              & Avg. Angular Tracking   & 0.992          & 0.991         &  & 0.836        & 1.004       & 0.991      &  & 0.956                                                                          & 0.992                  & 0.974                \\
                              & Total Episode Return    & 1.006          & 1.001         &  & 0.249        & 0.932       & 0.933      &  & 0.892                                                                          & 0.978                  & 0.908               \\ \Xhline{1.5pt} 
\end{tabular}
}
\end{table*}

\section{Experiments and Results}

We evaluate the effectiveness of \methodname{} in simulated environments, mainly focusing on three metrics: the average linear and angular velocity tracking return per step for task-related performance and the final total episode reward return for the overall locomotion quality. All the reported results are averaged over 5000 trajectories collected across all terrain types and levels and normalized over the performance of a pretrained privilege Oracle policy adopted from~\cite{kumar2021rma} on respective terrain, which is also used as the common teacher for all baselines for a fair comparison.

\subsection{Action Mixer Ratio}
\label{sec:mixer}
The first question we want to answer is: what is the optimal value for the Action Mixer introduced in Sec.~\ref{sec:action_mixer}? This mix ratio directly decides the proportion of the trajectories of the teacher and the student during the exploration and eventually affects the final performance due to the difference in their information density and the combined optimization objective with the next prediction and imitation of actin. To answer it, we train multiple \methodname{} models with different values of $\alpha$ and keep all other configurations untouched. Fig.~\ref{fig:mix_ratio} shows the key metrics for different terrains during testing.

We observe that the teacher policy suffers when the mix ratio is too low, as it cannot gather enough trajectories to reach an optimal policy, even with the help from all the privilege information. When the mix ratio is too high, the performance of the student starts to drop, as the training process is overly dependent on the guidance from the teacher policy and does not have enough exploration experience to handle out-of-the-distribution situations in complex environments. 

In most cases, the knowledge is transferred efficiently, and the student policy can achieve similar performance as its respective teacher. One special case is $\alpha=1$, where only the teacher trajectory is generated and used during training while the student is trained in a purely supervised manner. Although it produces one of the best performing teacher policies, the student acts poorly due to the lack of exploration, resulting in low survival rate. Another special case of $\alpha=0$ will be discussed in Sec.~\ref{sec:ablation} as it has a fundamental difference due to the lack of teacher participation and optimization.  

For the rest of this paper, we will use the policies trained with a ratio of $\alpha=0.6$ unless specified otherwise.

\subsection{Comparison with Baselines}
\label{sec:baseline_compare}

We compare \methodname{} with several knowledge transfer solutions and their variants with a base teacher network similar to~\cite{lai2023sim}:

\begin{itemize}
    \item \noindent\textbf{Supervised Transfer.} Based on~\cite{lee2020learning,kumar2021rma,lai2023sim}, we implemented different variants of supervised knowledge transfer with direct action imitation. \textbf{Offline-Only}: single stage with Oracle pre-trained policy used for trajectory generation~\cite{lee2020learning}; \textbf{Online-Only}: single stage with the student generating online trajectories with~\cite{kumar2021rma}; \textbf{Two-Stages}: combining two stages with offline pre-training first, followed by online correction~\cite{lai2023sim}.

    \item \noindent\textbf{Joint Transfer.} Following~\cite{radosavovic2023learning}, combining the RL exploration of the student with online supervised transfer with the joint ratio of policy imitation gradually annealed to zero by the mid-point of training.

    \item \noindent\textbf{CENet.} Implementation of an auto-encoder model from DreamWaQ~\cite{nahrendra2023dreamwaq} with VAE loss for the next observation, base velocity and latent space estimation with asymmetric actor-critic architecture.

    \item \noindent\textbf{PPO.} We use vanilla PPO~\cite{schulman2017proximal} to train \methodname{} solely on proprioceptive observation and RL loss, which is equivalent to an action mixer ratio of $\alpha=0$, and with the teacher head and other loss modules disabled, resulting in a student-only training.
    
\end{itemize}

The comparison results are summarized in Table~\ref{tab:comparison}. Although \methodname{} only uses the same number of trajectories as Oracle training, both the teacher and student policies achieve the similar performance level of Oracle performance and outperform other baseline models, which require many more trajectories to be generated for most cases. This shows the high efficiency of our proposed framework. With the increased difficulty of using omnidirectional control on all types of terrain, single-stage offline supervised transfer cannot efficiently capture the dynamics of the environment with just good trajectories, and the rough terrains make it hard for the student to survive. Although online supervised transfer and two-stage transfer significantly improve performance, \methodname{} still outperforms supervised transfer with fewer trajectories used.
Joint transfer require manual tuning of the joint ratio and struggles to handle the increased environmental challenges with similar performance to vanilla PPO with direct student-only training.

Learning with VAE loss shows strong performance, with only slight disadvantages compared to \methodname{}, making it one of the best performing baselines. These results strongly support the shared idea of DreamWaQ and \methodname{}, that understanding the state-action transition can significantly enhance policy optimization and overall performance. 

These results demonstrate that with our \methodname{} framework, optimal teacher and student policies can be achieved at the same time with a more compact unified network, without the need for multiple stages of knowledge transfer with complex network and loss function design. This greatly eases the training setup and the difficulty of knowledge transfer for quadruped locomotion for sim-to-real deployment.

\subsection{Ablation Studies}
\label{sec:ablation}

\begin{figure}[t]
    \centering
    \includegraphics[width=.95\columnwidth]{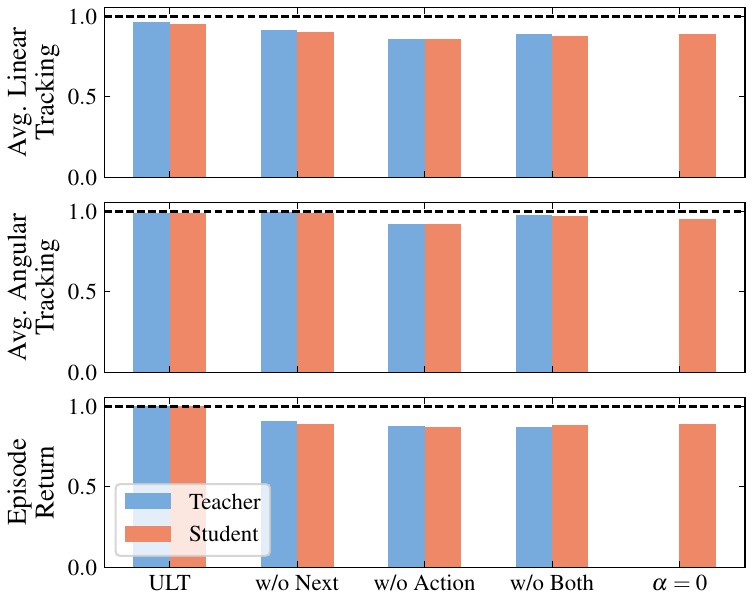}
    \caption{Normalized metrics for \methodname{} and its ablated variants. The return is averaged across trails on all five terrains.}
    \label{fig:ablation}
    \vspace{-10pt}
\end{figure}

To better evaluate the main components of \methodname{}, we create different variants by removing the next state-action predictions, imitation of actions, and both from the unified optimization pipeline. The results are shown in Fig.~\ref{fig:ablation}. With unified training, our core transformer benefits from both the next state-action prediction module and action imitation module, to increase its capability of understanding state-action transition with high-quality action guided by privilege information.

In addition, we explore the special case of the mix ratio $\alpha=0$, which is equipment to remove the action mixer module. As the teacher head will not be optimized, there is also no need to imitate the action. It is clear that missing the privilege information makes it difficult to optimize the next state-action prediction resulted in a performance similar to vanilla PPO. \methodname{} needs all modules to work together in order to achieve the sample and learning efficiency.

\begin{table}[t]
    \centering
    \caption{Performance of $\alpha=1$ student before and after supervised knowledge transfer.}
    \label{tab:transfer}
    \begin{tabular}{cccc}
    \Xhline{1.5pt}
    \textbf{} & \textbf{\begin{tabular}[c]{@{}c@{}}Avg. Linear\\ Tracking\end{tabular}} & \textbf{\begin{tabular}[c]{@{}c@{}}Avg. Angular\\ Tracking\end{tabular}} & \textbf{\begin{tabular}[c]{@{}c@{}}Episode\\ Return\end{tabular}} \\ \Xhline{1.5pt}
        Original  & 0.717 & 0.898 & 0.233 \\ 
        After Online Transfer  &         0.760  &     0.893                    &         0.865     \\ 
        \Xhline{1.5pt} 
    \end{tabular}
\end{table}

\subsection{\methodname{} with Supervised Transfer}
Training with proper mixed actions ensures that we can achieve optimal teacher and student policy at the same time, but a teacher-only policy still has slightly better performance. Can \methodname{} act as the teacher first and then as the classic supervised knowledge transfer in a single network? Tab.~\ref{tab:transfer} shows the average performance of the original student agent and after an online supervised transfer phase. Although we can recover some of the performance with additional transfer stage, it is still not comparable to \methodname{} as we can hardly update the transformer while fixing the output of the teacher head, demonstrating the importance and efficiency from the action mixer while training the framework as a whole.

\begin{figure*}[ht]
    \centering
    \includegraphics[width=.75\textwidth]{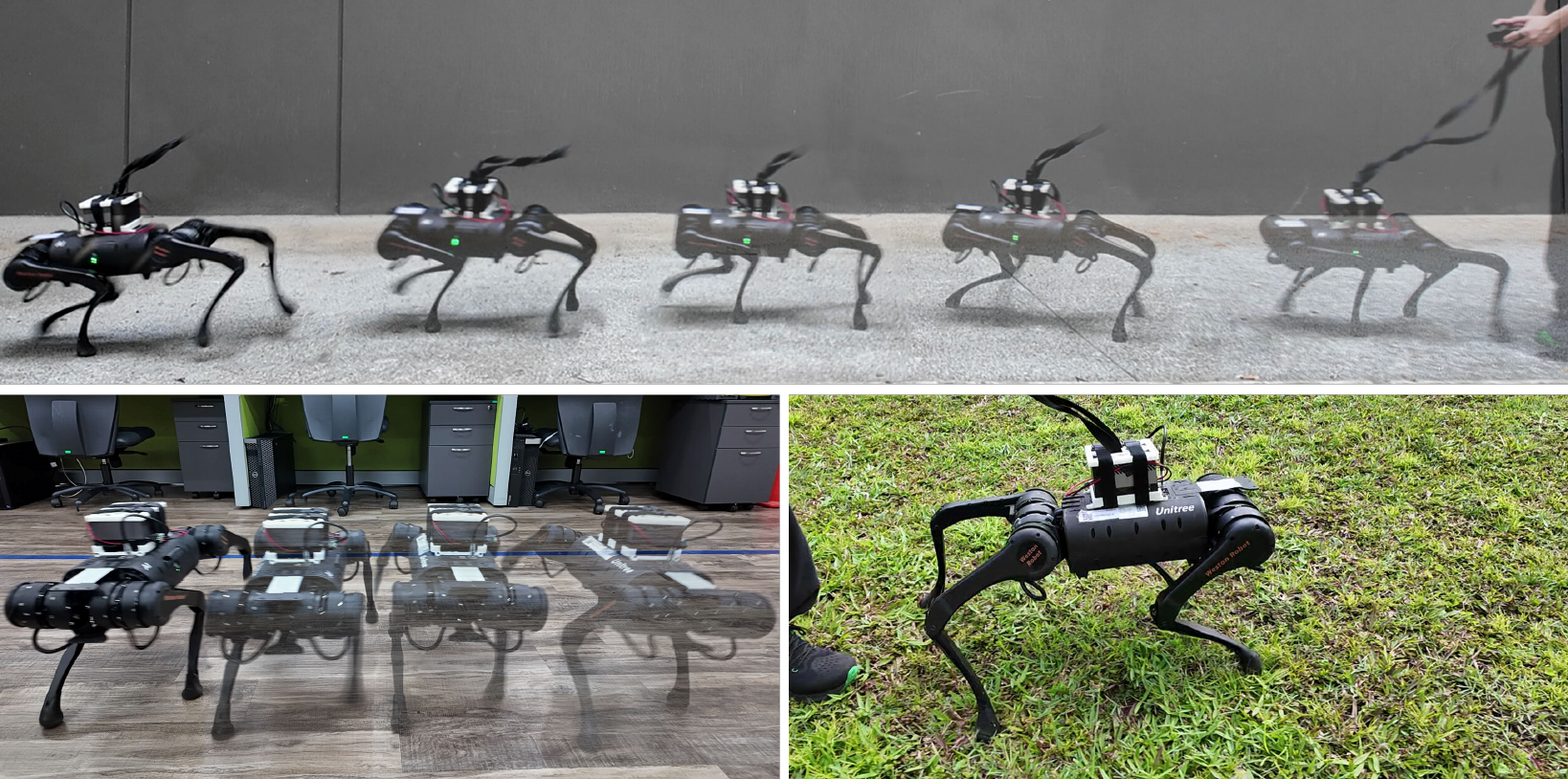}
    \caption{Footage snapshots showing a single trained \methodname{} policy deployed in the real world with zero-shot transfer on Unitree A1 equipped with Jetson Orin AGX for inference on different terrains with motion in omnidirections.}
    \label{fig:deployemnt}
    \vspace{-10pt}
\end{figure*}

\subsection{Physical Deployment}
Following Sector~\ref{sec:deploy}, we directly extract the \methodname{} framework and use onboard sensor observations to achieve zero-shot sim-to-real transfer. The policy is exported as JIT for portability and edge inference on a Unitree A1 robot equipped with a Jetson AGX Orin Developer Kit. The policy can run at up to 300Hz and we set the control frequency to 50Hz, $K_p = 30$ and $K_d = 0.7$ while communicating with A1's onboard low-level controller. Fig.~\ref{fig:deployemnt} shows some snapshots from the deployment test. Please refer to the supplementary video for more information.

\section{Conclusion}

We introduce \methodname{}, a unified framework based on transformers for simultaneous optimization of teacher and student policies for quadruped locomotion. With next state-action prediction and action imitation, \methodname{} can efficiently extract valuable transition information and provide guidance with privileged information for mixed exploration to improve the training process. This greatly reduces the complexity and trajectory data needed for sim-to-real transfer, enabling the direct deployment of the agent on physical systems. 

Although we have simplified the training process as one single phase without training a dedicated teacher policy, we still need to retrain the model when new task requirement is raised. It is appealing to explore the continual learning and generalization capability through the power of large language models (LLMs) in handling wide variety of information with multimodality and transfer to different tasks and embodiments as future work.

\bibliographystyle{IEEEtran}
\bibliography{references.bib}{}

\end{document}